# A.A. TELNYKH[1], I.V. NUIDEL[1], Yu.R. SAMORODOVA[2]

[1]Institute of Applied Physics RAS, Nizhny Novgorod

[2]Nizhny Novgorod State University, Nizhny Novgorod

telnykha@yahoo.com, nuidel@appl.sci-nnov.ru

# Construction of efficient detectors for character information recognition[*]

We have developed and tested in numerical experiments a universal approach to searching objects of a given type in captured video images (for example, people's faces, vehicles, special characters, numbers and letters, etc.). The novelty and versatility of this approach consists in a unique combination of the well-known methods ranging from creating detectors to making decisions independent of the type of recognition objects. The efficiencies of various types of basic features used for image coding, including the Haar features, the LBP features, and the modified Census transformation are compared. A combination of the modified methods is used for constructing 11 types of detectors of the number of railway carriages and for recognizing digits from zero to nine. The efficiency of the constructed detectors is studied.

Keywords: character information recognition, cascade detection, adaptive boosting, detector efficiency

## Paper structure

The main goal of the work is to construct effective tools for detecting numbers of railway carriages and digits. The study is based on known algorithms of recognition and detection of objects, including the adaptive boosting method for face recognition proposed by Viola and Jones [1,2], and methods for calculating the "convolutional" features of image fragments.

The Viola and Jones method [1,2] designed for detecting faces in images and the Adaboost method [3, 4] used to recognize symbolic information are briefly considered in the Introduction. In the sections to follow a combination of these methods modified to our tasks will be used for the development of detectors of various objects for character information recognition.

In the second section of the paper, the problem is formulated and the basic algorithms of the research, in particular, the modified Census transformation [5] are described. Why is the Census transformation taken? Since the system not only recognizes but, in the first place, detects an object, it is important to register features from various areas. The modified Census transform, that is a non-local binary pattern, is extended to a rectangle of arbitrary size inside the detection area. From the Viola-Jones recognition

---

[*] The work was supported by the Ministry of Education and Science of Russia (Project No. 14.Y26.31.0022)

method we borrow boosting, i.e. building weak and strong classifiers using the AdaBoost-learning method.

The numerical experiments, including number detection and digit recognition, building weak and strong classifiers using the AdaBoost method, as well as description of feature detectors are considered in the third section. The experiments were carried out on an example of railway carriage numbers.

In the fourth section we present the results of the numerical experiment aimed at comparing the efficiency of the detectors based on calculating (i) Census features [5] (non-local binary patterns), (ii) local binary patterns (LBP) [6, 7] successfully employed for recognizing textures, and (iii) Haar features [1, 2] used in the Viola-Jones algorithm for face recognition.

The conclusions are made in the fifth section.

## 1. Introduction

Video surveillance systems are currently widely used in various areas of human activity. Consequently, there is a high demand for intelligent video analytics systems – a technology using computer vision methods to automatically obtain diverse information based on the analysis of a sequence of images recorded with video cameras in real time or archived [8]. The interest in the modules of video analytics for the detection, recognition and tracking objects of symbolic information is extremely high. Typical objects of symbolic information are license plates of vehicles, railway car numbers, etc.

Optical recognition of any image is detecting and identifying an object or determining any of its properties by a recorded image [8,9]. The general algorithm for pattern recognition includes the following stages: 1) frame capture; 2) preprocessing; 3) localization of the object; 4) object recognition.

At the final stage, the area containing the object (or objects) must be recognized, i.e. attributed to one of the multitudes of classes, for example, character recognition in the detected license plate of the vehicle or identification of the detected person in the face recognition task.

In this work, emphasis is laid on character information recognition, namely, the recognition of license plates of railway carriages, i.e. capture of an object from the video flow and recognition of the moving symbolic objects.

There are various methods for solving the problem of moving symbolic objects recognition in video surveillance and recognition systems, such as template-based methods, methods using contour models, neural network methods [10,11], the Viola-Jones method [1,2], the Support Vector Method [12, 13], and others.

The Viola-Jones method of face recognition [1,2] is currently the basic technique of seeking objects in real time images, which has a low probability of false detection. This method is based on the following principles:

1) integral image representation; 2) Haar features; 3) boosting for selecting the most suitable features of the sought object in this part of the image; 4) all the features at the input of the binary classifier are sorted as "true" or "false"; 5) cascades of features quickly discard windows where the sought object (for example, a license plate) is not found.

AdaBoost selects a set of weak classifiers to be combined, then assigns its own weight to each. This weighted combination is a strong classifier. Viola and Jones combined the series of AdaBoost classifiers as a sequence of filters, which is especially effective for classifying image domains. Each filter is an independent AdaBoost classifier containing a rather small number of weak classifiers. The main drawback of the method is that the result of the work strongly depends on the training sample, since the light-sensitive brightness image acts as the input data.

The method of adaptive boosting [3, 4, 14] was used for recognizing traffic signs [15], number plates [16, 17] of motor vehicles and text images [18-21]. The authors of [22] considered image selection of railway carriage license plates by spatial frequencies using the Haar filters and the classification of the local region of search in the space of informative features with allowance for jump in brightness into two groups: object – not object. The problem of sign recognition was not solved. Number recognition by means of networks of deep learning was considered in [23, 24-28].

The analysis of the literature shows that detection of character information and identification of numbers based on a combination of the Viola-Jones and AdaBoost methods is attracting the attention of many researchers. The goal of the present work is to further develop the research started in the previous years [29-31].

## 2. Statement of the problem and algorithm description

We use the approach proposed by Viola and Jones [1,2] to detect numbers on railway cars and recognize the digits. As distinct from the features used in their work, the considered approach involves the so-called modified Census transform proposed in [5] and is extended to a rectangle of arbitrary size inside the detection area. The features are defined as a kernel 3x3 elements in size, in accordance with the spatial structure of the image. Within the kernel, binary {0.1} information encoding takes place, with the resulting binary patterns being boundaries, segments, saddle points, junction points, and so on, as is illustrated in fig. 1.

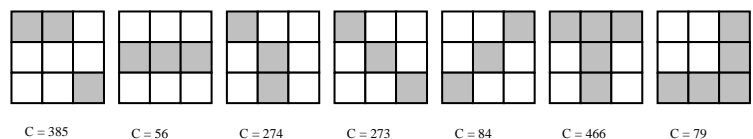

Fig. 1. An example of binary patterns used to encode information on the image.

The lattice of 3x3 elements contains $2^9$=512 such kernels. For encoding graphic information we use rectangular kernels of various sizes as shown in fig. 2.

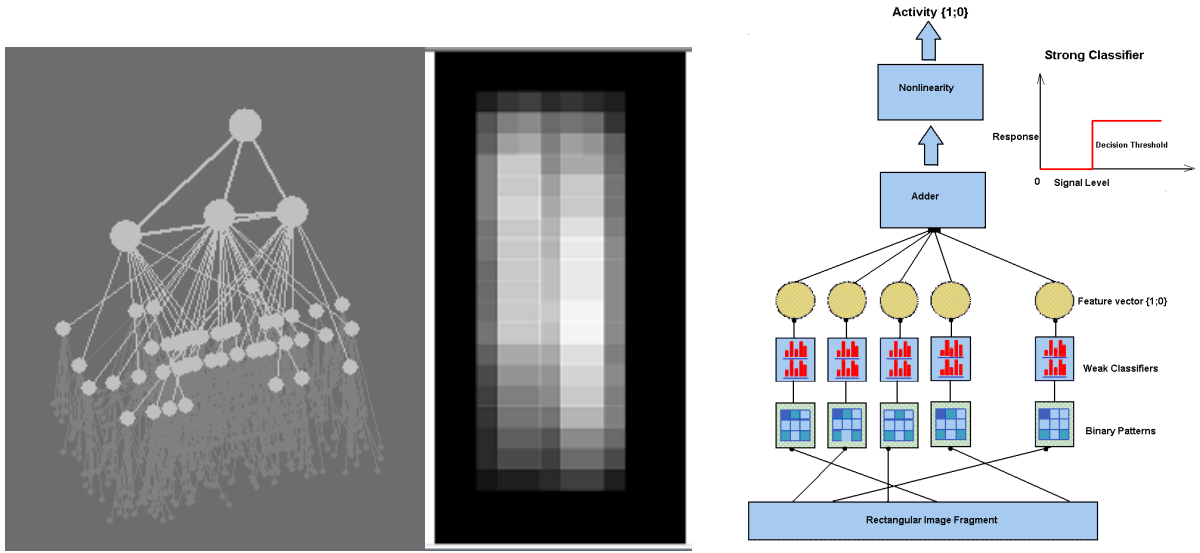

Fig. 2. a) 3D image of the network formed as a result of training; b) example of superposition of a rectangular 3x3 lattice within the aperture; c) scheme of a strong classifier.

Unlike the coding scheme in [5], the used coding scheme (1) does not allow image filtering. It transforms the image into a feature space with a dimension much larger than the size of the original image, as in the Viola and Jones scheme. We will call this kind of transformation a "non-local binary pattern". The formation of code description of the image is illustrated in fig.3.

The average brightness of a rectangular fragment of the image is determined as follows:

$$<I_F> = \frac{1}{w \cdot h} \sum_{x=x_o}^{x=x_0+w} \sum_{y=y}^{y=y_0+h} I(x,y) \quad (1),$$

where $I(x,y)$ is the brightness of the pixel with x, y-coordinates. The neighborhood of formation of code description is divided into 9 equal parts and the average brightness $<I_f>$ of each part is analyzed in comparison with the average brightness of the entire neighborhood of formation of the code description. The average brightness $<I_{f^n}>$ of the area belonging to the fragment is defined as

$$<I_{f^n}> = \frac{9}{w \cdot h} \sum_{x=x_i}^{x=x_i+w/3} \sum_{y=y_j}^{y=y_j+h/3} I(x,y) \quad (2),$$

$$c_{f^n} = \begin{cases} 1, <I_{f^n}> \geq <I_F> \\ 0, <I_{f^n}> < <I_F> \end{cases} \quad (3),$$

where $0 \leq i < 3, 0 \leq j < 3, n = i \cdot j$. For each domain, the brightness will be encoded according to the rule (3), where $c_{f^n}$ is the binary brightness code for each domain. In conformity with (3), for each image fragment there is a set of nine code bits $c_{f^0}.......c_{f^8}$. This sequence is considered to be code $C$ of any fragment of image I.

Thus, any rectangular fragment of the image is described by an integer number C having 9 bits width, i.e. $0 \leq C < 512$.

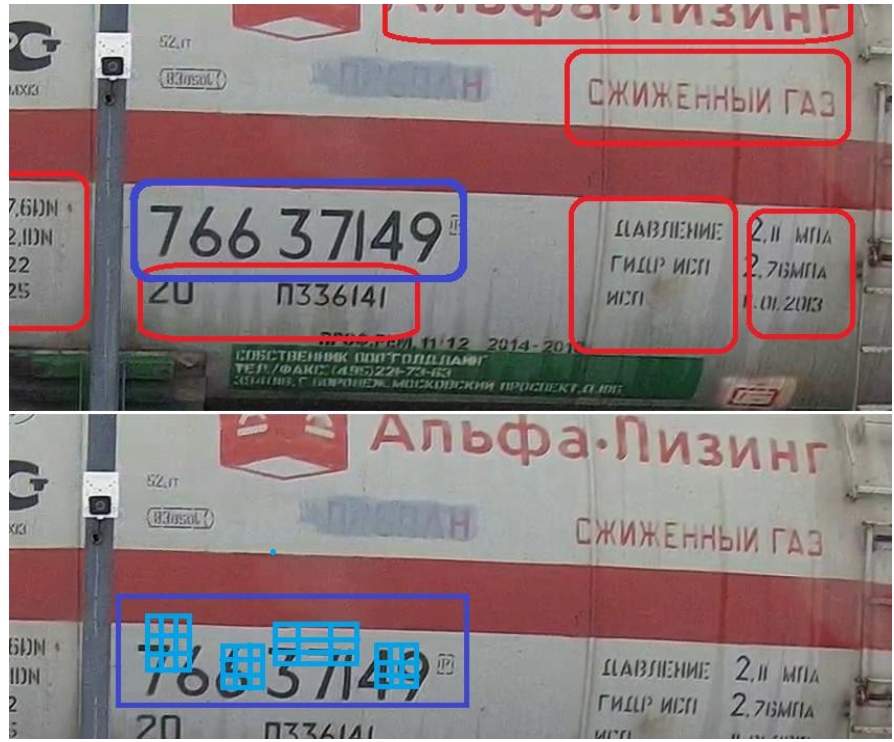

Fig. 3. Formation of code description of a rectangular section of binary image:
a) the considered image section with regions of diverse noises marked red and of the useful signal marked blue;
b) the region of detector aperture showing several variants of non-local binary patterns used for searching a given object in the image (car number).

### 3. Number detection and digit recognition, building weak and strong classifiers

#### 3.1. Weak classifier
On the basis of each feature, that is a non-local binary pattern, a weak classifier is created, which is a binary function taking on the values (0,1): zero in the absence of the sought object in the specified rectangular section of the image, and 1 otherwise (4):

$$h_k = \begin{cases} 1, \hat{L} > 0 \\ 0, \hat{L} = 0 \end{cases} \quad (4),$$

$$\hat{L} = \frac{\hat{\omega}_n(u'|\Omega=1)}{\hat{\omega}_n(u'|\Omega=0)} \quad (5),$$

where $L$ is the activation function of the weak classifier. To form the activation function, we used a decision rule based on the "maximum likelihood criterion" and its estimate (5), where $\hat{\omega}_n(u'|\Omega=1)$ is the probability density estimate of the input signal value obtained from the analysis of the training set, provided that the training sample represents samples of the "useful signal", and $\hat{\omega}_n(u'|\Omega=0)$ is the estimate of the probability density of the value of the input signal obtained by analyzing the training sample, provided that the training sample is a background sample.

Since the code has an integer nature, the estimate of its probability density distribution corresponds to the distribution histogram of the codes obtained from the analysis of the training sample.

### 3.2. Strong classifier

The so-called strong classifier (fig. 2c) is formed using the AdaBoost learning algorithm [3, 11, 12], a set of features, and weak classifiers. A strong classifier is a binary function {0,1} including a decision threshold determined in the learning process. The learning process minimizes recognition error in the training database:

$$H_i = \begin{cases} 1, \sum_{k=1}^{n} w_k h_k > \Theta \\ 0, \sum_{k=1}^{n} w_k h_k \leq \Theta \end{cases} \quad (6),$$

where $h_k$ is a weak classifier, $w_k$ is the weight of the weak classifier obtained in the learning process using the AdaBoost procedure, $n$ is the number of weak classifiers, and $\Theta$ is the decision making threshold.

### 3.3. Detector

The key element of the system of search and recognition of numbers on railway cars is the detector of objects of a given type. Objects in the image are detected using the cascade detection technique. The detector is a cascade of strong classifiers (fig. 2c) connected in series. For the search and recognition of the number of railway carriages, 11 types of detectors were created: a detector of a carriage number and a detector of all digits from 0 to 9. A general scheme of such a detector was described in [1, 2]. To compare the efficiency of different features, detectors with the use of Census features [5] were constructed as described above using binary patterns [6, 7] and Haar features [1,2].

## 4. Numerical experiment
### 4.1. Number detection

To detect the number, a database of images of the trains passing near a video camera installed at the railway station was collected. From the obtained videos, the frames showing the number of the railway carriage or tank car were selected. The total number of the selected frames was 1139 images. All the images were reduced to 640x480 pixels and discolored (represented in grayscale format). The area containing the number of the railway car was marked manually on each of the above images. The size of the minimum aperture of the detector was set to be 54x18 pixels. Then, the neighborhood of each marked area was exported to the training set as samples of the useful signal. The size of the sample depended on the degree of overlapping of the marked object and the scanning system of the detector.

Then the source database [36] was randomly divided into two parts with a ratio 3:1. The larger part was used for training the cascade structure of the detector (neural network). The smaller part did not participate in the training; it was used for finding the error level of the resulting detector. The size of the training sample was 865 images for training and 274 images for testing.

The recognition scheme described above was studied for different parameters. The first parameter is the degree of overlapping of image fragments to be recognized; the second parameter is the type of feature used: cs is a modified Census transform [5], LBP is a Local Binary Pattern [6, 7], haar means Haar features [1, 2]. The training and test samples were the same for all variants.

18 detectors were constructed and the characteristics of each of them were analyzed in the numerical experiments. The results of the numerical experiment on searching a number on a railway carriage are presented in figs. 4 and 5. The dependence of the number of features of different types on the size of the training sample is shown in fig. 4, and the dependence of the False Rejection Rate (FRR, type II errors) on the degree of overlapping and the type of the used feature in fig. 5. The False Acceptance Rate (FAR, type I errors) on the test sample was on the average ~ 5e-5 for each detector.

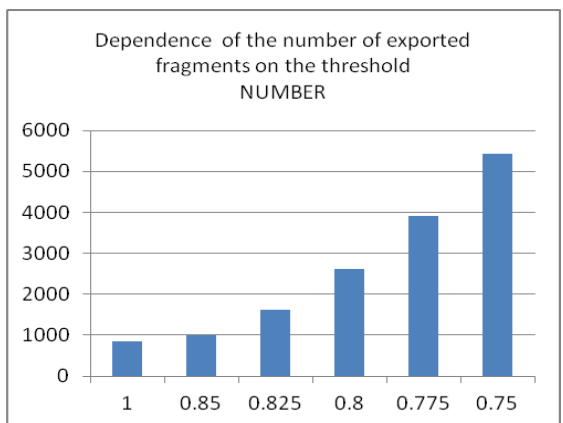
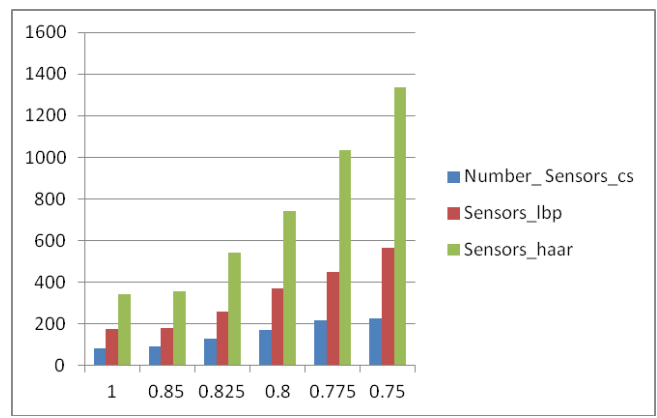

a) Number of fragments selected for training from the database [36] depending on overlap threshold.

b) Number of features (sensors) involved in the resultant detector depending on overlapping threshold and for different types of features (Haar, LBP and CS).

Fig. 4. Study of the size of neural network for detecting railway car number depending on overlapping threshold and type of features.

Analysis of the number of features used for achieving the FAR ~ 5e-5 shows that their number increases with increasing training sample size. The growth rate depends on the type of the used feature. The highest growth rate is for the Haar features [1], and the smallest for the modified Census features [5].

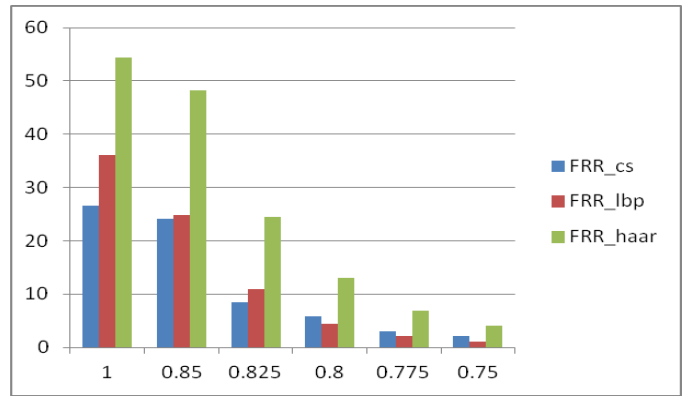

Fig. 5. The dependence of the FRR of a neural network for detecting railway car numbers on overlapping degree and type of feature used.

FRR analysis shows that the detectors constructed using the Haar features are significantly inferior to the detectors based on binary templates of various types (LBP and CS) not only in the number of elements used but also in the recognition quality.

**4.2. Digit Recognition**

A digit detector was trained using the database of the image numbers obtained. Each image was normalized to the size of 240x76 pixels and labeled by visible digits. The size of the digit detector aperture was 12x24 pixels. Then the neighborhoods of the image of each digit were exported to the database for training. As a result of the preparatory work, a database of the digits with the following characteristics was obtained (Table 1):

Table 1

| Digit | 0 | 1 | 2 | 3 | 4 | 5 | 6 | 7 | 8 | 9 |
|---|---|---|---|---|---|---|---|---|---|---|
| Training | 544 | 367 | 507 | 390 | 487 | 973 | 589 | 984 | 845 | 463 |
| Testing | 121 | 106 | 160 | 139 | 186 | 361 | 203 | 279 | 302 | 151 |
| FRR (%) | 2.48 | 31.13 | 1.25 | 2.88 | 3.76 | 2.77 | 7.39 | 4.84[*] | 2.32 | 0.66 |

The methodology of training the digit detector exactly corresponded to the methodology of teaching the number detector: the initial databases were divided with a 3:1 ratio into training and test parts. The detectors of each digit were built in the training part, depending on the degree of overlapping and the type of the used features. As a result, 180 digit detectors were obtained and their characteristics were analyzed. The result of the analysis of the digit detectors qualitatively coincides with the result of the analysis of the number detector: the growth rate of the number of features is different for different types of features.

It is higher for the Haar features and is minimal for the CS features, with the level of FAR and FRR errors higher for the Haar features than for the features built on binary patterns. The number of features used for different values of the overlapping degree and feature type is shown as an illustration in fig. 6. It should be noted that it is difficult to recognize some digits, in particular, distinguish 1 from 7, using this method, because they look alike.

In particular, digit "1" always occupies a smaller area than the other digits; therefore, the aperture of the 12x24 detector is large for it. In the course of training, fragments of other digits frequently get into it, and the resulting database, with overlapping taken into account, becomes unrepresentative. Therefore, for solution of this problem, additional study of the digit "1" using a smaller aperture (8x24) should be performed. It is also worthy of note that for most of the digits, the number of the used CS features is approximately equal to the number of the used features of the same type in a railroad car number detector.

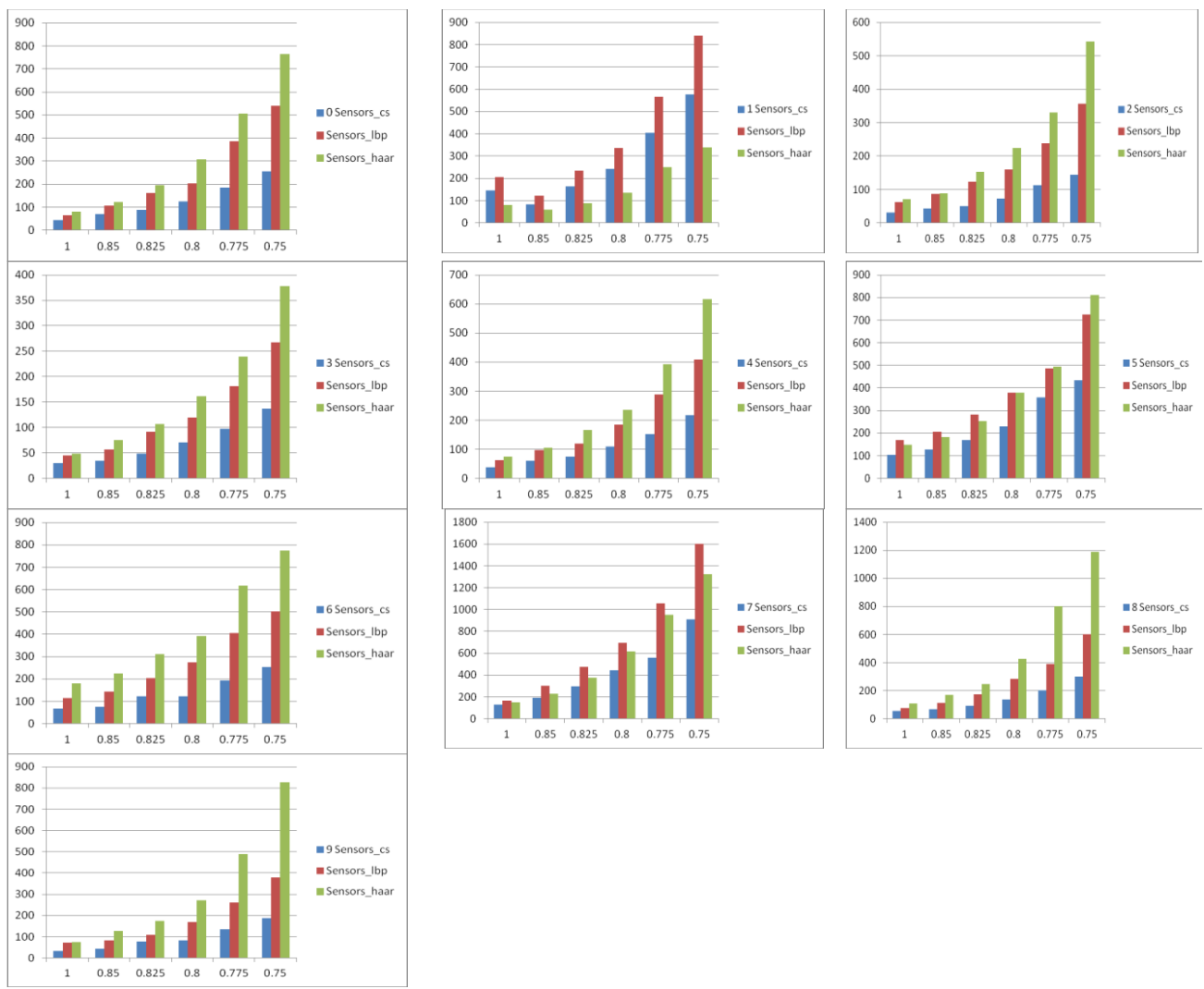

Fig. 6. Study of the size of the neural network for detecting digits on the railway car number depending on the overlapping threshold and type of the used features.

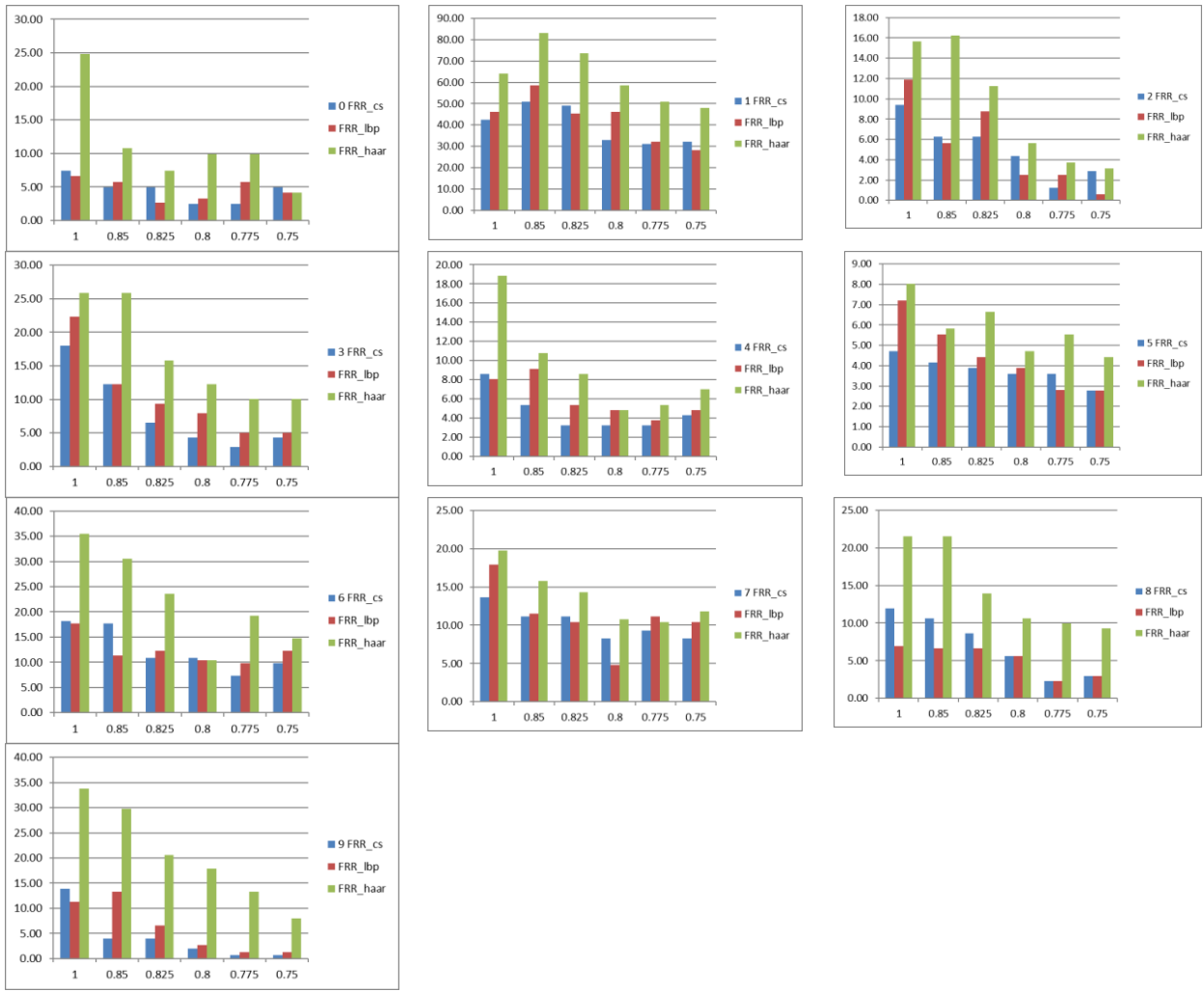

Fig. 7. The dependence of the FRR of neural network for detecting digits on the numbers of railway cars on the degree of overlapping and type of the used feature.

## 5. Conclusion

The development of efficient detectors for character information recognition on an example of recognizing license plates of railway cars has been considered, including capturing an object (number) from a video flow and recognizing objects. For detecting numbers and recognizing digits on railway cars, we have presented a modified general approach to constructing a cascade detector of objects of a given type proposed by Viola and Jones [1,2] and a comparative study of various options of basic features, including the Haar features [1,2], local binary patterns [6, 7] and the modified Census transformation [5]. We have proposed to use a combination of modified methods for constructing 11 types of detectors of various objects and have investigated their efficiency.

The features are modified so that a non-local rectangular binary pattern is formed, while the formal code description of the pattern is preserved. On the basis of each feature, a weak classifier is created, which is a binary function determined by the estimate of the distribution function for the useful signal and noise. The weak classifier is then used for the construction of a strong classifier following the AdaBoost algorithm.

The learning process minimizes recognition error on the training sample from the labeled database. The characteristics of 198 detectors constructed from a database [36] were analyzed depending on the overlapping parameter and the type of features used. It was shown that the growth rate of the number of features depends on their type, is maximal for the Haar features [1] and minimal for the feature constructed based on the modified Census transform. The FRR of the latter feature is minimal. It is shown that this conclusion is valid for both, number detection problem and recognition of the majority of the digits, except for digit 1.

On the basis of the results of numerical experiments we can say that for solving the problems of searching for objects of a given type, the best values of the parameter of overlapping of the labeled object and the scanning detector system lie in the [0.75, 0.8] range. The modified Census transform is most effective for forming the features of objects to be recognized.

### Further development of the research

1. We are planning to investigate the variation of the Census nuclei distribution depending on the objects to be recognized (number, 10 digits).

2. We intend to conduct qualitative and quantitative comparative analysis of the considered transformations and the transformations on convolutional neural networks (CNN) [23] with deep learning of the LeNet architecture for Caffe [35].

One of the arguments in favor of the developed system is briefly considered below. The organization of the system of learning and recognition of objects of any type in the developed system is biologically plausible (or neuromorphic). This is how the macrocolumns of the brain are organized: they consist of a network of neurons with receptive fields that respond to similar features [32-34]. The diameter of a microcolumn is 80-100 neurons located at five cellular levels. The microcolumns are combined into macromodules (or macrocolumns) $D = 500 - 100 \cdot 10^{-6} m$ each containing ~ 100 microcolumns. A typical artificial biomorphic model of a cortical column contains about 1000 neurons of different types and performs a strictly defined functional operation – the detection of objects of a given type.

The convolutional networks are models of receptive fields of neurons. However, in Deep Learning networks the network parameters are set initially, there is no object search, and only the recognition process takes place [23, 35].

Therefore, the developed system already has an advantage, as it is a biologically plausible model for searching objects (fig. 2a) which allows visualizing the layer-by-layer construction of the network in the learning process.

Qualitative and quantitative comparison of transformations is a separate issue which needs a deep further study.